\documentclass[runningheads]{llncs}
\usepackage{graphicx}
\usepackage{amsmath,amssymb} 
\usepackage{color}
\usepackage[ruled,vlined]{algorithm2e}
\usepackage[width=122mm,left=12mm,paperwidth=146mm,height=193mm,top=12mm,paperheight=217mm]{geometry}
\usepackage[normalem]{ulem}

\usepackage{caption}
\usepackage{subcaption}

\begin{document}
\pagestyle{headings}
\mainmatter
\def\ECCV18SubNumber{21}  

\title{Hard Occlusions in Visual Object Tracking} 

\titlerunning{Hard Occlusions in Visual Object Tracking}

\authorrunning{Thijs P. Kuipers, Devanshu Arya, Deepak K. Gupta}

\author{Thijs P. Kuipers, Devanshu Arya, Deepak K. Gupta}
\institute{Informatics Institute, University of Amsterdam, The Netherlands}

\maketitle

\begin{abstract}
Visual object tracking is among the hardest problems in computer vision, as trackers have to deal with many challenging circumstances such as illumination changes, fast motion, occlusion, among others. A tracker is assessed to be good or not based on its performance on the recent tracking datasets, \emph{e.g.}, VOT2019, and LaSOT. We argue that while the recent datasets contain large sets of annotated videos that to some extent provide a large bandwidth for training data, the hard scenarios such as occlusion and in-plane rotation are still underrepresented. For trackers to be brought closer to the real-world scenarios and deployed in safety-critical devices, even the rarest hard scenarios must be properly addressed. In this paper, we particularly focus on hard occlusion cases and benchmark the performance of recent state-of-the-art trackers (SOTA) on them. We created a small-scale dataset \footnote{Dataset can be accessed at https://github.com/ThijsKuipers1995/HTB2020} containing different categories within hard occlusions, on which the selected trackers are evaluated. Results show that hard occlusions remain a very challenging problem for SOTA trackers. Furthermore, it is observed that tracker performance varies wildly between different categories of hard occlusions, where a top-performing tracker on one category performs significantly worse on a different category. The varying nature of tracker performance based on specific categories suggests that the common tracker rankings using averaged single performance scores are not adequate to gauge tracker performance in real-world scenarios.

\end{abstract}

\section{Introduction}

Visual object tracking remains a challenging problem, even though it has been studied for several decades. A visual object tracker has to account for many different and varying circumstances. For instance, changes in the illumination may alter the appearance of the target object. The object could also blend in with the background environment, or it might get occluded, resulting in the object, or part of the object, being obstructed from view. Because of all the possible different circumstances, visual object trackers have to account for, visual tracking is considered a hard problem \cite{VT}.

Commonly, a tracker is assessed to be good or not based on its performance on established benchmarking datasets such as OTB100 \cite{OTB}, UAV123 \cite{UAV}, VOT2019 \cite{VOT2019}, GOT-10k \cite{GOT}, LaSOT \cite{lasot}, and TrackingNet \cite{TN}. These datasets comprise large sets of video sequences spanning across different challenges of tracking. For every dataset, the performance of a tracker is averaged over all sequences. 

For a tracker to be used in real-world scenarios and embedded in safety-critical equipment, it should be able to handle even the rarest and hardest instances of tracking. Therefore, the evaluation datasets should also contain such instances. We argue that this is not yet the case, and scenarios such as occlusion, in-plane rotation, and out-of-plane rotation are still underrepresented in these datasets. Moreover, most performance metrics compute an average score across all sequences, thereby overshadowing the poor performance of the subjected tracker on a certain specific challenge. For a deeper study on these aspects, we tailor the focus of this paper to only hard occlusions.

Occlusion refers to the phenomenon where parts of the target object are blocked from the field of view. For instance, an object can be occluded when it either moves partially or fully out of frame. Occlusion also occurs when another object fully or partially blocks the target object. When the target object is partially blocked, either certain specific features of the target object can disappear, or part of the entire target object appearance will disappear. Unlike other challenges of tracking, learning a distribution for occlusion is hard - no distribution exists in parts of the object that are occluded. This makes occlusion a hard problem. Some methods for handling occlusion do exist, but they are often focused on very specific aspects of tracking such as solely tracking pedestrians \cite{noh2018improving}. A major problem in evaluating visual object trackers on occlusion is the lack of data containing hard occlusions in current datasets. While the above-mentioned datasets do contain samples of occlusion, they often do not represent the challenging cases of occlusion that can occur when tracking in the wild \cite{OTB,TN,lasot}. Therefore, the available benchmarks might not accurately evaluate tracker performance on hard occlusions.

This work aims to evaluate a set of current state-of-the-art (SOTA) visual object trackers on the occurrences of hard occlusions. To perform this evaluation, a small dataset containing various samples of hard occlusions is compiled. Our preliminary results show that the performance of the SOTA trackers is, in general, lower on the hard occlusion scenarios. Further, we analyze whether the leading tracker among the ones chosen in this study performs superior on different scenarios of occlusion. Our results reveal interesting insights on whether even the best tracker, decided based on the existing evaluation strategies, could be considered a safe choice for deployment in real-world scenarios, especially for safety-critical devices. Further, it raises the question of whether the current performance metrics, averaging the score over all the sequences, are the right choice to assess the performance of trackers.

\section{Related Work}
In the following, we present an overview of some previous works that are relevant to this study. First, we present an overview of the recent visual object tracking algorithms, followed by works related to tackling occlusion in tracking. Finally, an overview of different tracking benchmarks is presented.

\subsection{Object Tracking}

The task of object tracking can refer to tracking multiple objects simultaneously (multi-object tracking) or tracking a single instance of an object (single-object tracking) \cite{TN}. This work will only consider \textit{generic} single-object trackers. To perform successful tracking, trackers must be able to create a strong appearance model for the target object and be capable of reliable and fast localization of the target object.

To address the above-mentioned challenge, various methods have been proposed. One such category is Correlation Filters (CF), which forms the basis of several SOTA single-object trackers \cite{TN}. CF uses circular correlation, which results in all shifted variants of input samples being implicitly included. This enables the construction of a powerful appearance model with very limited data \cite{DCF}. Furthermore, CF allows for fast run-times as computations are performed in the Fourier domain \cite{DCF3}. The MOSSE \cite{mosse} tracker paved the way for more advanced CF-based approaches such as the use of multi-dimensional features \cite{danelljan2014adaptive,multi,henriques2014high}, improved robustness to variations in scale and deformations \cite{li2014scale,danelljan2014accurate}, mitigating boundary effects \cite{danelljan2015learning,kiani2015correlation}, and the use of deep convolutional filters \cite{CCOT,ma2015hierarchical}. The advancements made in CF trackers resulted in large and complex models, which significantly increases the computational cost. ECO \cite{ECO} improves upon the CF-framework by reducing the model complexity by proposing a factorized convolution operation to increase running speeds.

Another category is deep learning-based trackers. Recurrent Neural Networks have been proposed for tracking \cite{gan2015first,kahou2017ratm}, but do not yield competitive performance compared to the SOTA trackers. MDNet \cite{nam2016learning} implements a deep convolutional network that is trained offline, and performs Stochastic Gradient Descent during tracking but is not able to operate in real-time. GOTURN \cite{held2016learning} utilizes a convolutional network to learn a function between image pairs. SiamFC \cite{FC} introduces a fully-connected Siamese architecture to perform general similarity learning. The goal is to learn a similarity function offline, which can then be evaluated during tracking to locate an exemplar image within a larger candidate image. SiamRPN \cite{RPN} implements a Siamese network that is extended with a Region Proposal Network (RPN) which allows for improved bounding-box predictions. SiamRPN++ \cite{RPN++} improves upon the Siamese architecture by enabling the use of deep networks in Siamese trackers. ATOM \cite{atom} improves upon the Siamese trackers by introducing an approach that consists of a target estimation module that is learned offline, and a target classification module that is learned online. DiMP \cite{dimp} consists of a discriminative model prediction architecture derived from a discriminative learning loss for visual tracking which can fully exploit background information.

\subsection{Occlusion in Tracking}

Occlusion remains largely unexplored in generic single-target object tracking. Some tracking architectures that do explicitly handle occlusion have been proposed. ROT \cite{ROT} utilizes occlusion-aware real-time object tracking by overcoming target model decay which can occur when the target object is being occluded.  In \cite{gupta2020tackling} SiamRPN++ and SiamFC are equipped with structured dropouts to handle occlusion. Other methods include more experimental strategies such as analyzing the occurrence of occlusion by utilizing spatiotemporal context information \cite{occ2}. By further analyzing motion constraints and the target reference the strategy allows for better discrimination between the target object and background distractors causing occlusion. Another experimental strategy uses layer-based strategies, extending it by specific background occluding layers \cite{layers}. Strategies that focus on handling occlusion have also been proposed in more specific object tracking tasks, such as tracking pedestrians \cite{noh2018improving} and cars \cite{car}. In \cite{multi} a multi-object tracker approach that handles occlusion is proposed, which is built on the idea of object permanence, using a region-level association and object-level localization process to handle long periods of occlusion. In \cite{lee2014occlusion} alternative SOTA methods for handling occlusion are presented such as depth-analysis \cite{greenhill2008occlusion,ma2010depth} and fusion methods such as a Kalman filter for predicting target object motion and location \cite{ali2009framework,zhao2009approach}. 

\subsection{Tracking Datasets} \label{datasets}

To evaluate the robustness of single-target visual object trackers, many datasets have been proposed. ALOV300 \cite{VT} contains 314 short sequences. ALOV300 does include 14 different attributes, including occlusion. However, it does not differentiate between different kinds of occlusion. OTB \cite{OTB} is another well-known dataset. The full dataset (OTB100) contains 98 sequences, while OTB50 is a subset of OTB100 containing 51 of the most challenging sequences. OTB offers 11 attributes, including both partial and full occlusion. Since the rise of deep trackers, the demand for large-scale datasets has increased. TrackingNet \cite{TN} is introduced to accommodate these demands. TrackingNet consists of over 30 thousand sequences with varying frame rates, resolutions, and lengths. TrackingNet includes 15 attributes, including both partial and full occlusion, as well as out-of-frame occlusion. GOT-10k \cite{GOT} consists of over 10 thousand sequences. GOT-10k offers an impressive 563 different object classes, and offers a train and test-set with zero overlaps between classes, resulting in more accurate evaluations. GOT-10k offers several attributes, including occlusion. Many of the sequences contained in the above-mentioned datasets are of relatively short duration. However, when tracking in the wild, tracking often occurs for sustained periods. LaSOT \cite{lasot} introduces a dataset consisting of 1400 longer duration sequences with an average sequence length of over 2500 frames. Similarly to the previously mentioned datasets, LaSOT includes a variety of attributes, including both partial and full occlusion, and out-of-frame occlusion. UAV123 \cite{UAV} contains 123 videos captured from a UAV, as well as artificial sequences generated by a flight-simulator.

\section{Benchmarking on Hard Occlusions}

\begin{figure}[h]
    \centering
    
    \begin{subfigure}{0.45\textwidth}
      \centering
      \includegraphics[width=\textwidth]{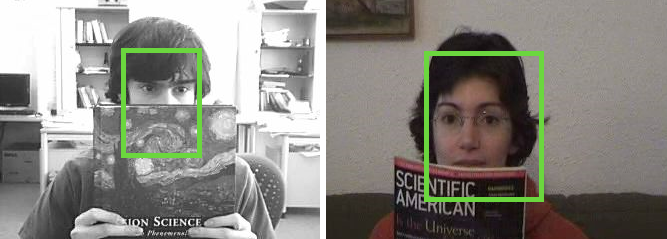}  
      \caption{Occlusion examples in OTB.}
      \label{fig:ex_a}
    \end{subfigure}
    \hfill
    \begin{subfigure}{0.45\textwidth}
      \centering
      \includegraphics[width=\textwidth]{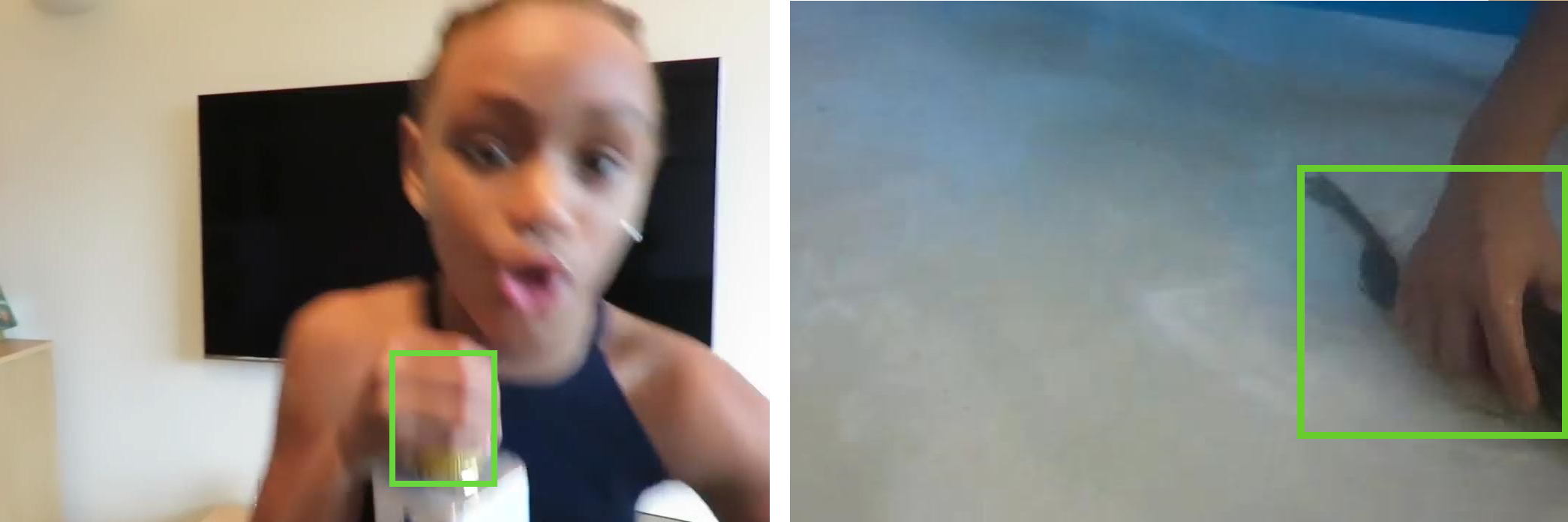}  
      \caption{Occlusion examples in LaSOT.}
      \label{fig:ex_b}
    \end{subfigure}
    
    \caption{Examples of occlusion from the OTB and LaSOT datasets.}
    \label{fig:occexamples}
\end{figure}

While most datasets described in Section \ref{datasets} contain cases of occlusion, the chosen instances are still very simple and do not account for the hard scenarios. These datasets do not take specific sub-categories within occlusion into account, as often solely general cases of occlusion, such as partial or full, are considered. Furthermore, many of the sequences contained in the datasets are of relatively short duration. As a result, occlusion often occurs for only short amounts of time. These short durations are not enough to accurately assess tracker performance on occlusion. Another issue is that often the occlusions that do occur involve simple cases. The occluded target object often possesses barely any movement relative to the camera, or the target object remains stationary throughout the sequence (see Figure \ref{fig:ex_a}). The challenging LaSOT \cite{lasot} dataset does contain more challenging cases of occlusion, including longer durations and more extreme movement of the target object (see Figure \ref{fig:ex_b}). However, the set of sequences containing these hard occlusions remains very limited.

\begin{figure}[h]
    \centering
    \includegraphics[width=10cm]{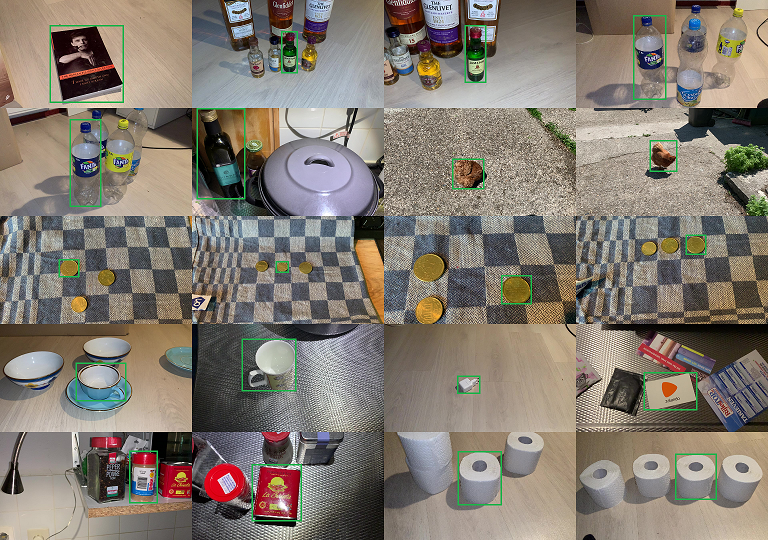}
    \caption{First frame of the sequences in HOB with ground-truth.}
    \label{fig:1st}
\end{figure}

Here, we present our Hard Occlusion Benchmark\footnote{The dataset can be accessed at https://github.com/ThijsKuipers1995/HTB2020} (HOB), a small-scale dataset containing 20 long-duration sequences that encompass a variety of different hard occlusion scenarios. For the sake of demonstration, Figure \ref{fig:1st} shows the first frame of some of the sequences with the corresponding ground-truth. Each sequence is of similar length, with an average of 2760 frames per sequence. Each sequence in HOB is annotated every 15th frame. Despite the lack of fully annotated sequences, with an average of 185 annotated frames per sequence, there exist ample ground-truths to perform an accurate evaluation. Naturally, HOB contains the general cases of hard occlusion, such as \textit{partial occlusion}, \textit{full occlusion}, and \textit{out-of-frame occlusion}. The cases of occlusion occur for long periods and are combined with strong movement and scale-variations of the target object relative to the camera. Also, these general cases are complemented with more specific attributes, to obtain a more precise evaluation of the SOTA tracker implementations on hard occlusions.

In its current form, HOB dataset comprises the following occlusion types.

\begin{itemize}
    \item \textit{Full out of frame occlusion (FOC).} 
    The target object moves fully out of the frame for extended periods. The target object may enter the frame at a different location compared to where it exited the frame.
    
    \item \textit{Feature occlusion (FO).}
    Some specific features of the target are omitted from view. It is still possible for the entire target object to be in view in this case. 
    
    \item \textit{Occlusion by transparent object (OCT).}
    The target object is being occluded by a transparent object. This means the target object can still be visible through the occluding object, although the occluding object does alter the appearance of the target object.
    
    \item \textit{Occlusion by similar object. (OCS)}
    The target object is being occluded by a similar-looking object.
\end{itemize}

\section{Experiments and Evaluations}
The current section introduces the set of trackers that are evaluated. Next, an overview of the metrics used for evaluation is given. Finally, the performance of the selected set of trackers is evaluated. The selected set of trackers aims to cover a variety of common state-of-the-art (SOTA) tracking principles. The following trackers are chosen for the evaluation: ECO \cite{ECO}, SiamFC \cite{FC}, SiamRPN++ \cite{RPN++}, ATOM \cite{atom}, DiMP \cite{dimp}. ECO proposes a general framework to improve upon the discriminant correlation filter tracking. SiamFC proposed similarity matching by using a fully connected Siamese network. SiamRPN++ uses a deep network for more sophisticated features, as well as a region proposal network. Three variants of SiamRPN++ are evaluated: the original version using the ResNet50 (r50) \cite{res} (r50) backbone, a version using the shallow AlexNet \cite{alex} backbone, and the long-term version which uses the long-term update strategy as described in \cite{VOT2018}. ATOM proposes a tracking architecture consisting of dedicated target estimation and target classification components. DiMP \cite{dimp} proposes a tracking architecture that can utilize both target and background information. Both ATOM and DiMP utilize a memory model to update the appearance model to take into account the change of appearance over time. Solely trackers with publicly available code are used in this work. 

\subsection{Evaluation Methodology}

We perform a One Pass Evaluation (OPE) measuring precision, success rate, area-under-curve, and the least-subsequence-metric \cite{LSM} on the 20 hard occlusion sequences in HOB. A brief overview of these metrics is presented below.

\subsubsection{Precision}

When tracking precision, the center localization error is calculated. The center localization error is defined as the Euclidean distance between the center of the ground-truth bounding box and the prediction bounding box. A frame is deemed successfully tracked if the calculated distance is below a certain threshold, say $t$. While precision does not take the predicted bounding box size into account, it does correctly measure how close the position of the prediction is to the ground truth, which is not always the case when using success rates, as only the overlap between prediction and ground truth is considered. In the case of occlusion, where the target object is not entirely visible, precision can depict to what extend the tracker manages to correctly predict the location of the occluded target object. The issue with using precision as a metric is its sensitiveness to resolution, which in the case of HOB is avoided since every sequence is of the same resolution. The final precision score for each tracker is evaluated using a threshold of 20 pixels such as in \cite{OBT}.

\subsubsection{Success Rate}

The success rate makes use of the average bounding box overlap, which measures performance by calculating the overlap between ground-truth bounding boxes and prediction bounding boxes. It takes both position accuracy and accuracy of the predicted bounding box size and shape into account. Therefore, the success rate can offer a solid measurement of tracker performance. Bounding box overlap is calculated using the intersection-over-union (IoU) score. Similar to precision, a frame is deemed successfully tracked when the calculated IoU meets a certain threshold, say $t$. By calculating the success rate at a range of different thresholds, a success plot can be formed. A threshold of $t > 0.5$ is often used to measure success. However, this does not necessarily represent a successfully tracked frame \cite{OTB}. Because of this, the area-under-curve (AuC) of the success plot is calculated instead, which takes the entire range of thresholds into account. Furthermore, frames in which the absence of the target object is correctly predicted are given an IoU score of 1. 

\subsubsection{Least Subsequence Metric}

The least-subsequence-metric (LSM) quantifies long term tracking behavior by computing the ratio between the length of the longest continuously ``successfully'' tracked sequence of frames and the full length of the sequence. A sequence of frames is deemed as successfully tracked if at least a certain percentage $p$ of frames within this is successfully tracked \cite{LSM}. The representative LSM score is calculated at a threshold of $p = 95\%$ as in \cite{LSM}. A frame is considered correctly tracked when the IoU of that frame is greater than $0.5$. Because LSM calculates the ratio between the longest continuously tracked subsequence and the length of the entire sequence, it can introduce a bias towards extremely long and short sequences. However, all sequences used in this work are of similar length, therefore this is not an issue for accurate evaluation.  

\subsection{Baseline Dataset}

For the sake of comparison with HOB, we use LaSOT as the baseline dataset. LaSOT is a large benchmark that focuses on long-term tracking, and it includes many sequences containing occlusions and out-of-frame occlusions. Due to this reason, it is one of the more difficult tracking benchmarks. Evaluating the selected visual object trackers on LaSOT will, therefore, offer a great baseline for comparing to HOB. HOB is a relatively small dataset containing only 20 sequences. To keep the comparison between HOB and LaSOT fair, only the top 20 occlusion heavy sequences from LaSOT are selected. Furthermore, while LaSOT offers per-frame annotations of ground-truths, HOB contains a ground-truth annotation every 15th frame. Therefore only every 15th frame of LaSOT will be used during the evaluation procedure.

\subsection{Overall Performance}

\begin{figure}[h]
    \centering
    \centerline{
        \includegraphics[width=\textwidth]{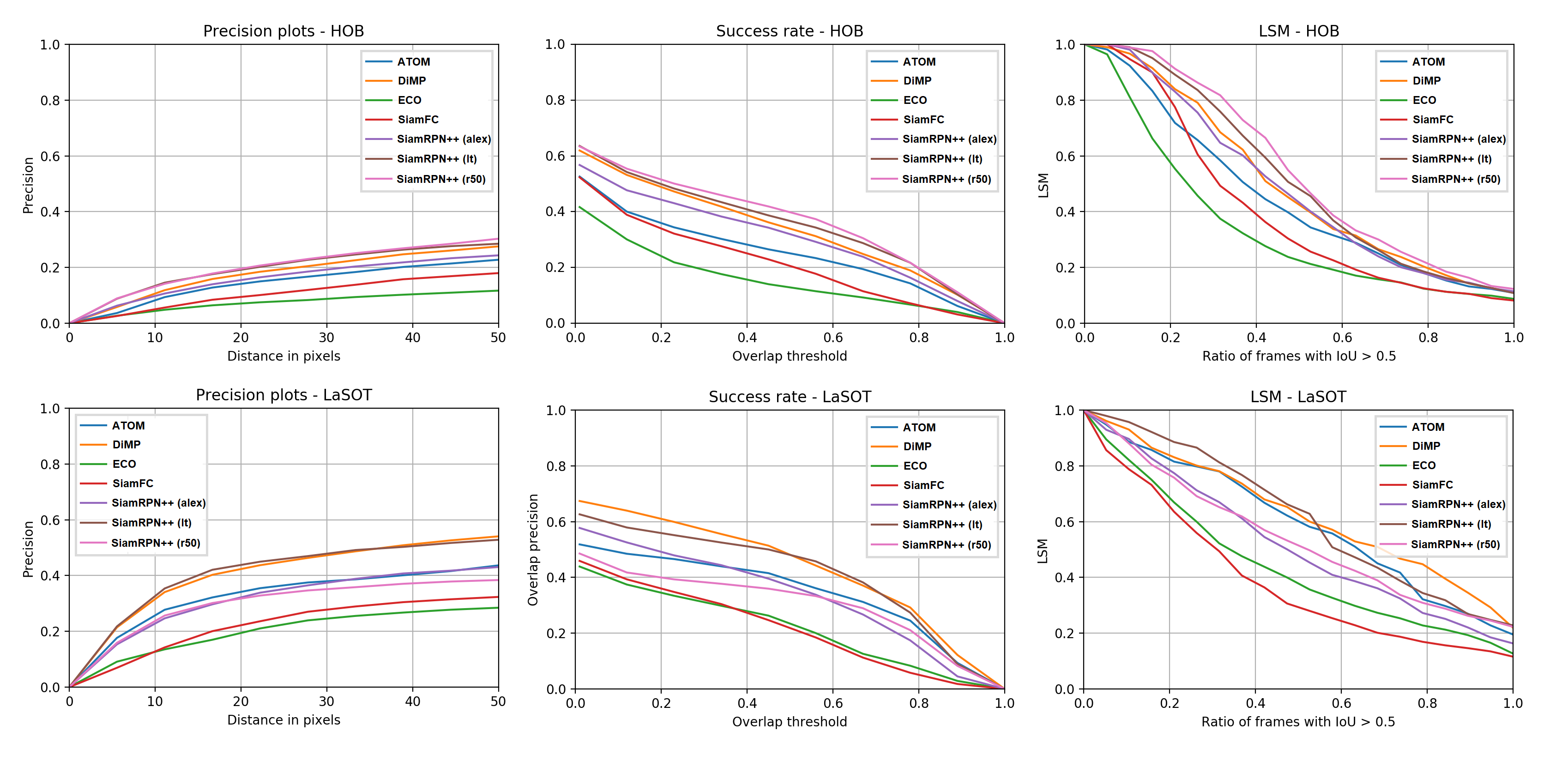}
    }
    \caption{Overall results on HOB (top) and LaSOT (bottom) on the precision rate, success rate, and LSM (left, middle, and right respectively).}
    \label{fig:overall}
\end{figure}

\ref{fig:attr}. Figure \ref{fig:predictions} depicts the predictions for each of the evaluated trackers on four sequences corresponding to the mentioned attributes.

\begin{table}[h]
\centering
\captionof{table}{Representative scores for precision (t $=$ 20), area under curve (AuC) and LSM (x $=$ 0.95) scores for each of the evaluated trackers on HOB and LaSOT. Best scores are shown in bold.}\label{table:hob}
\begin{tabular}{l|cc|cc|cl}
\textbf{}                 & \multicolumn{1}{r}{\textbf{Precision}} & \textbf{}      & \multicolumn{1}{r}{\textbf{AuC}}  & \textbf{}                           & \multicolumn{1}{r}{\textbf{LSM}} &                                    \\ \hline
\textbf{Dataset}          & \multicolumn{1}{c|}{\textbf{HOB}}      & \textbf{LaSOT} & \multicolumn{1}{c|}{\textbf{HOB}} & \multicolumn{1}{c|}{\textbf{LaSOT}} & \multicolumn{1}{c|}{\textbf{HOB}}                      & \multicolumn{1}{c}{\textbf{LaSOT}} \\ \hline
\textbf{ATOM}             & \multicolumn{1}{c|}{0.142}             & 0.342          & \multicolumn{1}{c|}{0.243}        & \multicolumn{1}{c|}{0.317}          & \multicolumn{1}{c|}{0.122}                             & 0.228                              \\
\textbf{DiMP}             & \multicolumn{1}{c|}{0.173}             & 0.421          & \multicolumn{1}{c|}{0.324}        & \multicolumn{1}{c|}{\textbf{0.391}}          & \multicolumn{1}{c|}{0.126}                             & \textbf{0.292}                              \\
\textbf{ECO}              & \multicolumn{1}{c|}{0.070}             & 0.191          & \multicolumn{1}{c|}{0.149}        & \multicolumn{1}{c|}{0.199}          & \multicolumn{1}{c|}{0.098}                             & 0.166                              \\
\textbf{SiamFC}           & \multicolumn{1}{c|}{0.093}             & 0.225          & \multicolumn{1}{c|}{0.205}        & \multicolumn{1}{c|}{0.191}          & \multicolumn{1}{c|}{0.090}                             & 0.135                              \\
\textbf{SiamRPN++ (alex)} & \multicolumn{1}{c|}{0.154}             & 0.320          & \multicolumn{1}{c|}{0.300}        & \multicolumn{1}{c|}{0.300}          & \multicolumn{1}{c|}{0.125}                             & 0.185                              \\
\textbf{SiamRPN++ (lt)}   & \multicolumn{1}{c|}{0.192}             & \textbf{0.437}          & \multicolumn{1}{c|}{0.343}        & \multicolumn{1}{c|}{0.383}          & \multicolumn{1}{c|}{0.127}                             & 0.248                              \\
\textbf{SiamRPN++ (r50)}  & \multicolumn{1}{c|}{\textbf{0.195}}             & 0.318          & \multicolumn{1}{c|}{\textbf{0.359}}        & \multicolumn{1}{c|}{0.278}          & \multicolumn{1}{c|}{\textbf{0.133}}                             & 0.245                             
\end{tabular}
\end{table}

Figure \ref{fig:overall} shows the precision, success rate, and LSM of each evaluated tracker on both HOB and LaSOT. In Table \ref{table:hob}, the representative scores of each of the metrics are shown. The results show that on average, the performance of the evaluated trackers is worse on HOB compared to LaSOT. SiamRPN++(r50) is the top-performing tracker on HOB on all metrics. SiamRPN++(r50) outperforms SiamRPN++(lt) by a small margin on the AuC metric, and similar observations can be made for the other two metrics as well. This is an interesting result, as SiamRPN++(lt) is specifically tailored to handling long-term tracking which includes occlusion of the target object. These results imply that even the re-detection module of SiamRPN++(lt) can occasionally drift the tracker model to false targets. This could be attributed to SiamRPN++(lt) re-detecting the wrong target object and sticking to it during long and heavy stretches of occlusion, which would result in lower overall performance. Contrary to the results obtained on HOB, performance on LaSOT seems significantly different. On LaSOT, DiMP is the top-performing tracker on the AuC and LSM metrics, and second-best on precision. Only SiamRPN++(lt) shows comparative performance, and as shown in Table \ref{table:hob}. The performance of the remaining trackers is significantly lower.

It is interesting to note that DiMP consistently underperforms compared to both SiamRPN++(r50) and SiamRPN++(lt) on HOB. While HOB dataset contains only 20 sequences, the sequences of HOB have been chosen with no intended bias towards Siamese and SiamRPN-type trackers. Thus, we would ideally expect DiMP to perform the best, as has been seen on LaSOT. One possible reason for the reduced performance on HOB could be attributed to the fact that the training set of DiMP comprised the training set of LaSOT sequences as well. This could mean that DiMP tends to overfit on objects similar to those observed in the LaSOT training set. Since most other datasets contain similar tracking instances, DiMP performs well on them. On the contrary, the scenarios observed in HOB are quite different, and that leads to reduced performance on this dataset. Another possible reason for the performance decay of DiMP on HOB could be attributed to the bias added to the tracking model due to frequent model updates happening even under the scenarios of occlusion \cite{Gavves2019arxiv}. This is not the case for the SiamRPN++ variants, as they do not perform any model update. Note that Table \ref{table:hob} shows a relatively large difference in precision scores between HOB and LaSOT compared to the AuC scores. This is partly caused by the lower resolution sequences of LaSOT, as precision is sensitive to resolution.

On HOB, ECO is the worst performing tracker in terms of, precision and AuC. On LaSOT, ECO and SiamFC are the worst-performing trackers, with ECO obtaining a slightly higher AuC score. It seems that the discriminative correlation filter approach utilized in ECO is not very well suited for occlusions and long-term tracking in general, as it may not be able to generalize compared to the Siamese based trackers. In the case of SiamFC, its lack of accurate target classification and localization capabilities seems to hamper performance during cases of hard occlusion. This becomes more apparent in the LSM score, where SiamFC and ECO are the lowest-performing trackers, as a low LSM score indicates frequent loss of the target object. ATOM performs consistently worse on HOB compared to the three SiamRPN++ variants. On LaSOT, ATOM performs very similarly to SiamRPN++(r50) and SiamRPN++(alex), generally outperforming them by a slight margin. ATOM also utilizes a model update strategy, which could result in the decay of the appearance model during cases of hard occlusion.

\subsection{Attribute Evaluation}

\begin{figure}[h]
    \centering
    \includegraphics[width=\textwidth]{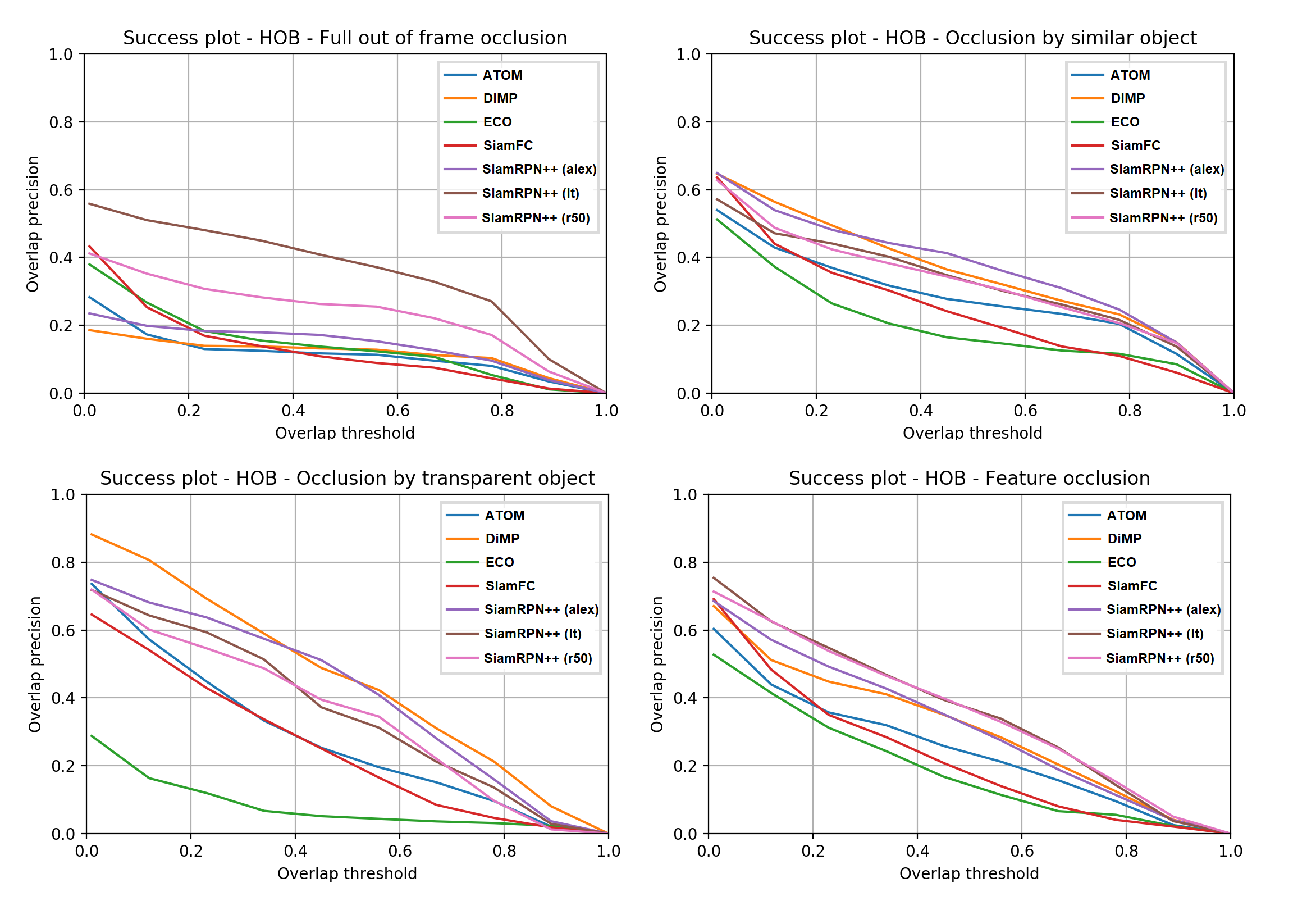}
    \caption{Success plots for full out of frame occlusion (top left), occlusion by similar object (top right), occlusion by transparent object (bottom left), and feature occlusion (bottom right).}
    \label{fig:attr}
\end{figure}

\begin{figure}
    \centering
    
    \begin{subfigure}{\textwidth}
      \centering
      \includegraphics[width=\textwidth]{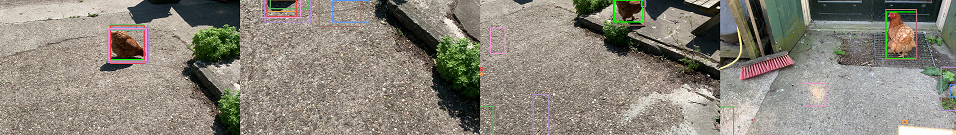}  
      \caption{Full out of frame occlusion.}
      \label{fig:foc}
    \end{subfigure}
    
    \begin{subfigure}{\textwidth}
      \centering
      \includegraphics[width=\textwidth]{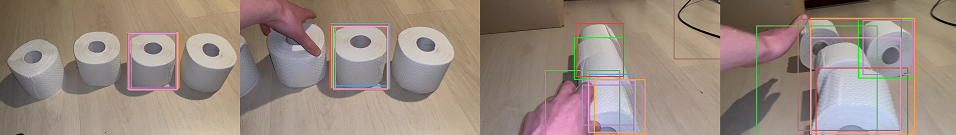}  
      \caption{Occlusion by similar object.}
      \label{fig:ocs}
    \end{subfigure}
    
    \begin{subfigure}{\textwidth}
      \centering
      \includegraphics[width=\textwidth]{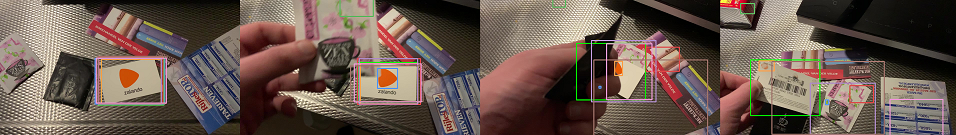}  
      \caption{Feature occlusion.}
      \label{fig:oct}
    \end{subfigure}
    
    \begin{subfigure}{\textwidth}
      \centering
      \includegraphics[width=\textwidth]{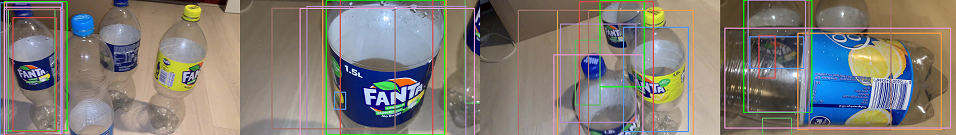}  
      \caption{Occlusion by transparent object.}
      \label{fig:fo}
    \end{subfigure}
    
    \begin{subfigure}{\textwidth}
      \centering
      \includegraphics[width=\textwidth]{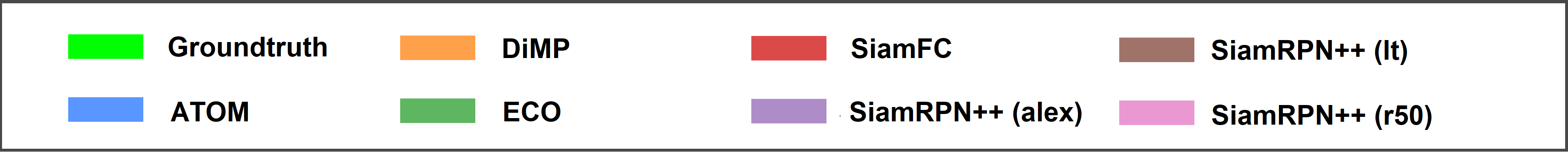}  
      \label{fig:sub-first}
    \end{subfigure}
    
    \caption{Depiction of four frames including prediction and groundtruth bounding-boxes for each the categories \textit{full out-of-frame occlusion}, \textit{occlusion by similar object}, \textit{feature occlusion}, and \textit{occlusion by transparent object}.}
    \label{fig:predictions}
\end{figure}

While the results from the previous section have shown that most trackers struggle in the presence of hard occlusions, it is of interest to analyze further how different occlusion types affect the overall performance. In this section, we study the trackers for the different categories of occlusion that we have defined earlier. The success plots for each of the categories are shown in Figure \ref{fig:attr}. Figure \ref{fig:predictions} depicts the predictions for each of the evaluated trackers on sequences corresponding to each of the categories. 

\textit{Full out-of-frame occlusion (FOC).} FOC seems to be a very challenging problem for the visual object trackers, with SiamRPN++(lt) being the top-performing tracker in this category. This is most likely attributed to its re-initialization strategy when detecting target object loss. The second-best performing tracker on FOC is SiamRPN++(r50), performing considerably better compared to the rest. Having access to rich features at different scales seems to aid its re-detection capabilities when the target object moves within the localization area. When observing the predictions, only SiamRPN++(lt) seems to consistently be able to re-detect the target object (see Figure \ref{fig:foc}). Overall, the SiamRPN++ variants outperform the other evaluated trackers. DiMP performs considerably worse during FOC. Interestingly, even trackers with weak discriminative power, such as SiamFC and ECO, perform on par with DiMP and ATOM for cases of FOC. ATOM, DiMP, and ECO update their appearance model during tracking. In the case of FOC, this could result in the appearance updating on samples that do not contain the target object causing strong appearance model decay. This is not the case for the Siamese trackers, as their appearance model remains fixed during tracking.

\textit{Occlusion by similar object (OCS).} In the case of OCS, SiamRPN++(alex) has the highest overall performance, while SiamRPN++(r50) performs the worst of the SiamRPN++ trackers. Interestingly, the use of the shallow AlexNet as a backbone results in better performance compared to using the deep ResNet, even outperforming the long-term SiamRPN++ variant. Re-initialising on target loss does not offer an advantage during OCS, as the performance of SiamRPN++(lt) is similar to the performance of SiamRPN++(alex). DiMP is the second-best performing tracker, with ATOM, SiamFC and, ECO being the lowest-performing trackers. ECO performs considerably lower compared to the other trackers. When observing the predictions during OCS, trackers struggle to accurately keep track of the target object (see Figure \ref{fig:ocs}).

\textit{Feature occlusion (FO).} During FO, SiamRPN++(r50) and SiamRPN++(lt) are the top performing trackers, with near-identical performance. As objects tend to stay at least partially visible in this category, the re-initialization strategy of SiamRPN++(lt) does not offer much benefit in tracking performance. SiamRPN++(lt) and SiamRPN++(r50) are closely followed by SiamRPN++-(alex) and DiMP. Once again, ECO is the worst performing tracker. The results of the FO category are very similar to the overall performance on occlusion as shown in Figure \ref{fig:overall}, although on average the trackers seem to perform slightly worse on feature occlusion specifically at higher thresholds.

\textit{Occlusion by transparent object (OCT).} DiMP is the best performing tracker on OCT, SiamRPN++(alex) being a very close second. Both DiMP and Siam-RPN++(alex) perform considerably better than to the other SiamRPN++ variants, similar to the OCS category. SiamRPN++(r50) and SiamRPN++(lt) have very similar performance. In the cases of OCS and OCT, it seems that the ability to generalize for objects that are not seen during training could play an important role. The appearance model predictor implemented in DiMP contains few parameters, leading to better generalization as less overfitting to observed classes occurs during the offline training phase \cite{dimp}. Likewise, SiamRPN++(alex) using AlexNet contains less parameters compared to SiamRPN++(r50) \cite{RPN++}. The performance of ATOM is considerably lower compared to DiMP, while both use the same IoU maximization based architecture for updating the appearance model, suggesting the appearance model update is of less importance during OCT and OCS. It is interesting to note that when observing the predictions on OCT and FE, DiMP and ATOM tend to strongly focus on striking target object features, as can be observed in \mbox{Figure \ref{fig:oct}} and \mbox{Figure \ref{fig:fo}}.

\section{Conclusion}

In this work, we presented an evaluation of the current state-of-the-art (SOTA) visual object trackers on hard occlusions. We compiled a small dataset containing sequences that encompass several samples of hard occlusions to assess the performance of these trackers in such occluded scenarios. Furthermore, we evaluated the trackers on a subset of the most occlusion-heavy LaSOT sequences. From the results, we show that on average the trackers perform worse on hard occlusion scenarios, suggesting that occlusion is still a relatively unsolved problem in tracking. While DiMP is the best performing tracker on the LaSOT benchmark, it is consistently outperformed by SiamRPN++ using the ResNet backbone architecture (r50) and its long-term tracking variant (lt) on instances of hard occlusions. Furthermore, we show that the top-performing tracker can vary drastically between specific scenarios of hard occlusions. For example, while DiMP seems the best for handling occlusions caused by semi-transparent objects, it performs the worst for full out-of-frame occlusion scenarios.

The set of results presented in this paper hint towards the fact that even the best performing tracker based on the current benchmark datasets might not be suited for real-world deployment, especially in safety-critical applications, such as self-driving cars. Real-world problems do not promise the presence of a uniform set of challenges, and at any random instance, a different tracking challenge could be the most important. Correspondingly, we focused on the challenge of hard occlusions in this paper, and trackers behaved differently than they did on LaSOT. This implies two important things for future research. First, tracking datasets need to incorporate more instances of difficult tracking challenges. Second, evaluation methodologies need to be designed that give more importance to instances where a certain tracker performs the worst. To summarize on a high-level, a model that handles even the most difficult challenges of tracking sufficiently well should be considered a better visual object tracker.

\bibliographystyle{splncs}
\bibliography{egbib}

\begin{thebibliography}{10}

\bibitem{VT}
Smeulders, A.W., Chu, D.M., Cucchiara, R., Calderara, S., Dehghan, A., Shah,
  M.:
\newblock Visual tracking: An experimental survey.
\newblock IEEE transactions on pattern analysis and machine intelligence
  \textbf{36}(7) (2013)  1442--1468

\bibitem{OTB}
Wu, Y., Lim, J., Yang, M.H.:
\newblock Object tracking benchmark.
\newblock IEEE Transactions on Pattern Analysis and Machine Intelligence
  \textbf{37} (09 2015)  1--1

\bibitem{UAV}
Mueller, M., Smith, N., Ghanem, B.:
\newblock A benchmark and simulator for uav tracking.
\newblock In: European conference on computer vision, Springer (2016)  445--461

\bibitem{VOT2019}
Kristan, M., Matas, J., Leonardis, A., Felsberg, M., Pflugfelder, R.,
  Kamarainen, J.K., \v{C}ehovin Zajc, L., Drbohlav, O., Lukezic, A., Berg, A.,
  Eldesokey, A., Kapyla, J., Fernandez, G.:
\newblock The seventh visual object tracking vot2019 challenge results (2019)

\bibitem{GOT}
Huang, L., Zhao, X., Huang, K.:
\newblock Got-10k: A large high-diversity benchmark for generic object tracking
  in the wild.
\newblock IEEE Transactions on Pattern Analysis and Machine Intelligence (2019)

\bibitem{lasot}
Fan, H., Lin, L., Yang, F., Chu, P., Deng, G., Yu, S., Bai, H., Xu, Y., Liao,
  C., Ling, H.:
\newblock Lasot: A high-quality benchmark for large-scale single object
  tracking.
\newblock In: Proceedings of the IEEE conference on computer vision and pattern
  recognition. (2019)  5374--5383

\bibitem{TN}
Muller, M., Bibi, A., Giancola, S., Alsubaihi, S., Ghanem, B.:
\newblock Trackingnet: A large-scale dataset and benchmark for object tracking
  in the wild.
\newblock In: Proceedings of the European Conference on Computer Vision (ECCV).
  (2018)  300--317

\bibitem{noh2018improving}
Noh, J., Lee, S., Kim, B., Kim, G.:
\newblock Improving occlusion and hard negative handling for single-stage
  pedestrian detectors.
\newblock In: Proceedings of the IEEE Conference on Computer Vision and Pattern
  Recognition. (2018)  966--974

\bibitem{DCF}
Danelljan, M., Hager, G., Shahbaz~Khan, F., Felsberg, M.:
\newblock Learning spatially regularized correlation filters for visual
  tracking.
\newblock In: Proceedings of the IEEE international conference on computer
  vision. (2015)  4310--4318

\bibitem{DCF3}
Bolme, D.S., Beveridge, J.R., Draper, B.A., Lui, Y.M.:
\newblock Visual object tracking using adaptive correlation filters.
\newblock In: 2010 IEEE computer society conference on computer vision and
  pattern recognition, IEEE (2010)  2544--2550

\bibitem{mosse}
{Bolme}, D.S., {Beveridge}, J.R., {Draper}, B.A., {Lui}, Y.M.:
\newblock Visual object tracking using adaptive correlation filters (2010)

\bibitem{danelljan2014adaptive}
Danelljan, M., Shahbaz~Khan, F., Felsberg, M., Van~de Weijer, J.:
\newblock Adaptive color attributes for real-time visual tracking.
\newblock In: Proceedings of the IEEE Conference on Computer Vision and Pattern
  Recognition. (2014)  1090--1097

\bibitem{multi}
Huang, Y., Essa, I.:
\newblock Tracking multiple objects through occlusions.
\newblock In: 2005 IEEE Computer Society Conference on Computer Vision and
  Pattern Recognition (CVPR'05). Volume~2., IEEE (2005)  1051--1058

\bibitem{henriques2014high}
Henriques, J.F., Caseiro, R., Martins, P., Batista, J.:
\newblock High-speed tracking with kernelized correlation filters.
\newblock IEEE transactions on pattern analysis and machine intelligence
  \textbf{37}(3) (2014)  583--596

\bibitem{li2014scale}
Li, Y., Zhu, J.:
\newblock A scale adaptive kernel correlation filter tracker with feature
  integration.
\newblock In: European conference on computer vision, Springer (2014)  254--265

\bibitem{danelljan2014accurate}
Danelljan, M., H{\"a}ger, G., Khan, F., Felsberg, M.:
\newblock Accurate scale estimation for robust visual tracking.
\newblock In: British Machine Vision Conference, Nottingham, September 1-5,
  2014, BMVA Press (2014)

\bibitem{danelljan2015learning}
Danelljan, M., Hager, G., Shahbaz~Khan, F., Felsberg, M.:
\newblock Learning spatially regularized correlation filters for visual
  tracking.
\newblock In: Proceedings of the IEEE international conference on computer
  vision. (2015)  4310--4318

\bibitem{kiani2015correlation}
Kiani~Galoogahi, H., Sim, T., Lucey, S.:
\newblock Correlation filters with limited boundaries.
\newblock In: Proceedings of the IEEE Conference on Computer Vision and Pattern
  Recognition. (2015)  4630--4638

\bibitem{CCOT}
Danelljan, M., Robinson, A., Khan, F.S., Felsberg, M.:
\newblock Beyond correlation filters: Learning continuous convolution operators
  for visual tracking.
\newblock In: European conference on computer vision, Springer (2016)  472--488

\bibitem{ma2015hierarchical}
Ma, C., Huang, J.B., Yang, X., Yang, M.H.:
\newblock Hierarchical convolutional features for visual tracking.
\newblock In: Proceedings of the IEEE international conference on computer
  vision. (2015)  3074--3082

\bibitem{ECO}
Danelljan, M., Bhat, G., Shahbaz~Khan, F., Felsberg, M.:
\newblock Eco: Efficient convolution operators for tracking.
\newblock In: Proceedings of the IEEE conference on computer vision and pattern
  recognition. (2017)  6638--6646

\bibitem{gan2015first}
Gan, Q., Guo, Q., Zhang, Z., Cho, K.:
\newblock First step toward model-free, anonymous object tracking with
  recurrent neural networks.
\newblock arXiv preprint arXiv:1511.06425 (2015)

\bibitem{kahou2017ratm}
Kahou, S.E., Michalski, V., Memisevic, R., Pal, C., Vincent, P.:
\newblock Ratm: recurrent attentive tracking model.
\newblock In: 2017 IEEE Conference on Computer Vision and Pattern Recognition
  Workshops (CVPRW), IEEE (2017)  1613--1622

\bibitem{nam2016learning}
Nam, H., Han, B.:
\newblock Learning multi-domain convolutional neural networks for visual
  tracking.
\newblock In: Proceedings of the IEEE conference on computer vision and pattern
  recognition. (2016)  4293--4302

\bibitem{held2016learning}
Held, D., Thrun, S., Savarese, S.:
\newblock Learning to track at 100 fps with deep regression networks.
\newblock In: European Conference on Computer Vision, Springer (2016)  749--765

\bibitem{FC}
Bertinetto, L., Valmadre, J., Henriques, J.F., Vedaldi, A., Torr, P.H.:
\newblock Fully-convolutional siamese networks for object tracking.
\newblock In: European conference on computer vision, Springer (2016)  850--865

\bibitem{RPN}
Li, B., Yan, J., Wu, W., Zhu, Z., Hu, X.:
\newblock High performance visual tracking with siamese region proposal
  network.
\newblock In: Proceedings of the IEEE Conference on Computer Vision and Pattern
  Recognition. (2018)  8971--8980

\bibitem{RPN++}
Li, B., Wu, W., Wang, Q., Zhang, F., Xing, J., Yan, J.:
\newblock Siamrpn++: Evolution of siamese visual tracking with very deep
  networks.
\newblock In: Proceedings of the IEEE Conference on Computer Vision and Pattern
  Recognition. (2019)  4282--4291

\bibitem{atom}
Danelljan, M., Bhat, G., Khan, F.S., Felsberg, M.:
\newblock Atom: Accurate tracking by overlap maximization.
\newblock In: Proceedings of the IEEE Conference on Computer Vision and Pattern
  Recognition. (2019)  4660--4669

\bibitem{dimp}
Bhat, G., Danelljan, M., Gool, L.V., Timofte, R.:
\newblock Learning discriminative model prediction for tracking (2019)

\bibitem{ROT}
Yilmaz, A., Li, X., Shah, M.:
\newblock Contour-based object tracking with occlusion handling in video
  acquired using mobile cameras.
\newblock IEEE Transactions on pattern analysis and machine intelligence
  \textbf{26}(11) (2004)  1531--1536

\bibitem{gupta2020tackling}
Gupta, D.K., Gavves, E., Smeulders, A.W.:
\newblock Tackling occlusion in siamese tracking with structured dropouts.
\newblock arXiv:2006.16571 (2020)

\bibitem{occ2}
Pan, J., Hu, B.:
\newblock Robust occlusion handling in object tracking.
\newblock In: 2007 IEEE Conference on Computer Vision and Pattern Recognition,
  IEEE (2007)  1--8

\bibitem{layers}
Zhou, Y., Tao, H.:
\newblock A background layer model for object tracking through occlusion.
\newblock In: Proceedings Ninth IEEE International Conference on Computer
  Vision, IEEE (2003)  1079--1085

\bibitem{car}
Koller, D., Weber, J., Malik, J.:
\newblock Robust multiple car tracking with occlusion reasoning.
\newblock In: European Conference on Computer Vision, Springer (1994)  189--196

\bibitem{lee2014occlusion}
Lee, B.Y., Liew, L.H., Cheah, W.S., Wang, Y.C.:
\newblock Occlusion handling in videos object tracking: A survey.
\newblock In: IOP conference series: earth and environmental science.
  Volume~18., IOP Publishing (2014)  012020

\bibitem{greenhill2008occlusion}
Greenhill, D., Renno, J., Orwell, J., Jones, G.A.:
\newblock Occlusion analysis: Learning and utilising depth maps in object
  tracking.
\newblock Image and Vision Computing \textbf{26}(3) (2008)  430--441

\bibitem{ma2010depth}
Ma, Y., Chen, Q.:
\newblock Depth assisted occlusion handling in video object tracking.
\newblock In: International Symposium on Visual Computing, Springer (2010)
  449--460

\bibitem{ali2009framework}
Ali, A., Terada, K.:
\newblock A framework for human tracking using kalman filter and fast mean
  shift algorithms.
\newblock In: 2009 IEEE 12th International Conference on Computer Vision
  Workshops, ICCV Workshops, IEEE (2009)  1028--1033

\bibitem{zhao2009approach}
Zhao, J., Qiao, W., Men, G.Z.:
\newblock An approach based on mean shift and kalman filter for target tracking
  under occlusion.
\newblock In: 2009 International Conference on Machine Learning and
  Cybernetics. Volume~4., IEEE (2009)  2058--2062

\bibitem{res}
He, K., Zhang, X., Ren, S., Sun, J.:
\newblock Deep residual learning for image recognition.
\newblock In: Proceedings of the IEEE conference on computer vision and pattern
  recognition. (2016)  770--778

\bibitem{alex}
Krizhevsky, A., Sutskever, I., Hinton, G.E.:
\newblock Imagenet classification with deep convolutional neural networks.
\newblock In: Advances in neural information processing systems. (2012)
  1097--1105

\bibitem{VOT2018}
Kristan, M., Leonardis, A., Matas, J., Felsberg, M., Pflugfelder, R.,
  Cehovin~Zajc, L., Vojir, T., Bhat, G., Lukezic, A., Eldesokey, A.,  et~al.:
\newblock The sixth visual object tracking vot2018 challenge results.
\newblock In: Proceedings of the European Conference on Computer Vision (ECCV).
  (2018)  0--0

\bibitem{LSM}
Moudgil, A., Gandhi, V.:
\newblock Long-term visual object tracking benchmark.
\newblock arXiv preprint arXiv:1712.01358 (2017)

\bibitem{OBT}
{Wu}, Y., {Lim}, J., {Yang}, M.:
\newblock Online object tracking: A benchmark.
\newblock In: 2013 IEEE Conference on Computer Vision and Pattern Recognition.
  (2013)  2411--2418

\bibitem{Gavves2019arxiv}
Gavves, E., Tao, R., Gupta, D.K., Smeulders, A.W.M.:
\newblock Model decay in long-term tracking.
\newblock arXiv:1908.01603 (2019)

\end{thebibliography}
\end{document}